# Deep Learning Features at Scale for Visual Place Recognition

Zetao Chen, Adam Jacobson, Niko Sünderhauf, Ben Upcroft, Lingqiao Liu, Chunhua Shen, Ian Reid
and Michael Milford[1]

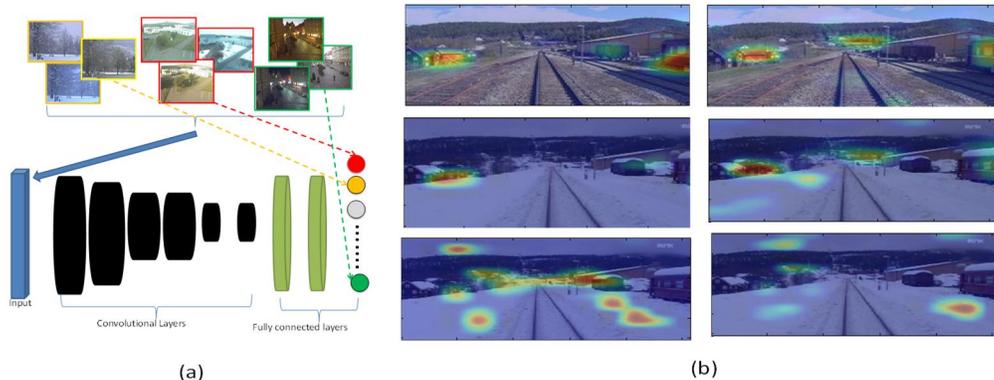

Figure 1 (a) We have developed a massive (2.5M image) place recognition-specific dataset containing 1000s of places with hundreds of exemplars of each place under changing conditions, enabling the training of two CNN models to learn condition-invariant features for place recognition across extreme appearance conditions. Training images collected from the same camera are assigned to the same label in the final layer (such as red, yellow and green dots in Figure 1(a)) (b) Feature heat maps overlapped on top of the testing images. Each column demonstrates an example of successful place recognition across extreme weather conditions achieved using this approach. Two images from Spring (top row) are localized against images in the Winter (middle and bottom rows). Features generated by our model (middle row) fire at semantically meaningful locations, even in the presence of strong appearance changes. Features generated by the ImageNet model (bottom row) generate location responses that are less meaningful when environmental conditions change, likely because this is a scenario not explicitly encountered in its training dataset.

*Abstract*— The success of deep learning techniques in the computer vision domain has triggered a range of initial investigations into their utility for visual place recognition, all using generic features from networks that were trained for other types of recognition tasks. In this paper, we train, at large scale, two CNN architectures for the specific place recognition task and employ a multi-scale feature encoding method to generate condition- and viewpoint-invariant features. To enable this training to occur, we have developed a massive Specific PlacEs Dataset (SPED) with hundreds of examples of place appearance change at thousands of different places, as opposed to the semantic place type datasets currently available. This new dataset enables us to set up a training regime that interprets place recognition as a classification problem. We comprehensively evaluate our trained networks on several challenging benchmark place recognition datasets and demonstrate that they achieve an average 10% increase in performance over other place recognition algorithms and pre-trained CNNs. By analyzing the network responses and their differences from pre-trained networks, we provide insights into what a network learns when training for place recognition, and what these results signify for future research in this area.

## I. INTRODUCTION

Place recognition can be considered as an image retrieval task which consists of determining a match between the current scene and a previously visited location. State-of-the-art visual place recognition algorithms such as FAB-MAP [1] match the appearance of the current scene to a previously visited place by converting the image into bag-of-words representations [2] built on local features such as SIFT or SURF. However, recent evidence suggests that features extracted from Convolutional Neural Networks (CNNs) trained on very large datasets significantly outperform SIFT features on a variety of vision tasks [3], such as object recognition [4], fine-grained recognition [5], scene recognition [6] and object detection [7].

Motivated by these results, recent studies have also shown that state-of-the-art performance in place recognition can be achieved by utilizing intermediate representations from CNNs that have already been trained on object recognition datasets [8-10]. McManus et al. [11] used an SVM to learn patch-based distinctive visual elements, called scene signatures, to match scenes under appearance changes. Their approach yielded excellent performance but has the highly restricting requirement that training must occur in the *test* environment under all possible environmental conditions. Finally, networks have been trained to recognize *types* of places (scenes) [12] rather than *specific* places. However, the task of scene recognition is different from place recognition; images under the same scene category can come from different places. *Specific* place recognition is the key component of loop closure in the vast majority of mapping and localization systems.

Consequently it is clear there are several major gaps in our knowledge and capability regarding deep learning and place recognition: we do not know what training deep networks specifically for the task of place recognition will yield; we do not have place recognition datasets of sufficient scale with which to properly investigate this question; and we do not understand the characteristics of networks trained specifically for place recognition and how they are different to networks trained for other tasks.

[1] ZC is with the Vision for Robotics Lab at ETH Zurich, chenze@ethz.ch. AJ, NS, BU and MM are with the Australian Centre for Robotic Vision at the Queensland University of Technology. LL, CS and IR are with the Australian Centre for Robotic Vision at the University of Adelaide. This work was supported in part by an Australian Research Council Future Fellowship FT140101229 to MM and an ARC Centre of Excellence grant to MM ,BU, CS and IR. This work was done when ZC was in the Australian Centre for Robotic Vision.

In this paper we address these issues, presenting several advances towards the training of deep networks specifically for place recognition performance, at scale. To enable this training to occur, we have developed a large (2.5 million) image place recognition dataset containing thousands of places and hundreds of examples of each place under a wide range of environmental conditions, differentiating this training dataset in scale and coverage from any previous one [13]. The multiple exemplars of each place enables us to cast the place recognition problem as a classification problem. Using this new dataset, we evaluate the trained network's place recognition performance on four publicly available benchmark place recognition datasets and provide comparisons to conventional algorithmic place recognition techniques and approaches built on top of networks trained for other recognition tasks. Finally, by visualizing the CNN layers' responses, we illustrate the differences in the internal representation of a network trained specifically for place recognition versus other recognition tasks. We make both the datasets and the trained place recognition network freely available online upon publication.

The paper is organized as follows. Section II provides an overview of visual place recognition and datasets used in this area. In Section III we describe how we train the network and encode the deep learning features. Section IV we illustrate how we construct the dataset. Experiments are presented in Section V, with results shown in Section VI. Finally we conclude the paper in Section VII.

## II. BACKGROUND

In this section we briefly review previous work utilizing CNNs for place recognition and currently available place recognition datasets.

### A. Visual Place Recognition with Convolutional Networks

Visual place recognition over perceptually-changing environments generally falls into two categories: utilizing feature representations that are robust to perceptual changes [11, 14-16] or learning and predicting appearance changes [17-20].

These methods either operate directly on the raw image pixels or rely on a fixed set of traditional hand-crafted features. However, it is rapidly becoming apparent in the computer vision community that hand-crafted features are being outperformed by deep learnt features in various vision tasks [3-7], which prompts the question of whether we can learn better features automatically for place recognition.

[8] was the first work to introduce a CNN-based place recognition system and [9] provided a thorough investigation of the utility and viewpoint-invariant properties of deep learnt features for place recognition. In both [10] and [21], the authors combined a landmark proposal technique with convolutional neural network features to match patches over extreme appearance and viewpoint variations. However, these studies all utilize pre-trained CNNs that are trained on an object-centric dataset, which is different in nature from the place recognition task. The question is: can training a CNN specifically for place recognition further improve place recognition performance?

[22] trained a CNN for a different but relevant task of camera pose estimation and [13] was the first to train a CNN particularly for place recognition. It appears that the absolute performance figures achieved on datasets shared in common with this research were significantly poorer (less than 40% maximum recall on the Nordland dataset), possibly due to the challenges of training a CNN without a sufficiently large and varied dataset, a challenge we address here by the creation of the SPED dataset. [23] also proposed to train a CNN for place recognition. Different from their work that interpreted place recognition as a triplet matching process, we formulate place recognition as a classification task which can be solved more efficiently on our SPED dataset with millions of images.

### B. Existing Place Recognition Datasets

Several benchmarks for place recognition have been constructed and studied in the literature.

The Eynsham dataset is a large 70 km road-based dataset (2 × 35 km traverses) used in the FAB-MAP [24] and SeqSLAM studies [16, 25]. The whole dataset contains 9575 panoramic images captured at 7 meter intervals. The St. Lucia dataset [26] comprises images recorded from a selection of streets at 5 different times of a day over a period of two weeks. It only captures images in suburb environments and the appearance variations of each place is small. The Nordland dataset [27] consists of a total 10 hours of video footage covering a 728 km journey four times, once in each season. It is a perfect dataset for studying season changes in natural environments. However, this dataset only captures seasonal changes and does not contain enough variations in illumination, day-night cycles or other weather phenomena.

All of these place recognition datasets are tiny in comparison with current object datasets such as ImageNet, and do not offer more than a handful of exemplars of what each place looks like under varying environmental conditions. In order to train a generally-applicable condition-invariant CNN for place recognition, a much larger dataset is required, one that captures a wide range of condition variations in a much wider variety of environments. The closest existing dataset is the scene-centric "places" database constructed in [12] – but scene recognition ("kitchen") is fundamentally different to recognizing a specific and unique place ("the location of the kitchen on the

4th floor of the Burns building").

We construct and make freely available for the first time a large-scale place-centric database which consists of images captured from surveillance cameras around the world. The cameras were selected so that the dataset contains images obtained in a wide variety of environments. As of now, this dataset contains more than 2.5 million images. The details of the dataset are described in Section IV.

## III. TRAINING CNNS FOR PLACE RECOGNITION

In this section we describe how we train CNNs specifically for place recognition using the new dataset described in Section IV, as well as how we encode the deep learning features for improving their viewpoint robustness.

We consider three different network architectures: a classification architecture [28], a Siamese network [29] and a triplet network [30]. Because there are millions of images in the SPED dataset, training such a huge dataset on a Siamese or triplet network will generate exponentially large permutation, leading to impractical training time. On the other hand, the classification architecture has been shown to successfully train a network on millions of images [28]. The SPED dataset contains thousands of exemplars of each single place which can be better interpreted as a classification problem – that is, classifying each image to its correct place rather than comparing pairs or triplets of images. As a result, we train a classification network for this task.

### A. Training CNN for Place Recognition

To train the CNN, we randomly select 1,272,000 images from the constructed database as the training set, with an average of approximately 500 images from each camera and a total of 2543 cameras. The validation set contains about 50 images per camera, resulting in a total of 120,000 images.

In this task, let $\{I_i, y_i\} \in R^N \times C$ denotes the $i^{th}$ image where $I_i \in R^N$ and $y_i$ is the corresponding place label with $C$ different possible values. Our classification ConvNet contains six convolutional layers (with max-pooling) to learn and extract features hierarchically, followed by two fully-connected layer to learn more complex non-linear mapping, and a softmax output layer in the end indicating place identity. All training images are RGB images and are all resized to $256 \times 256$. The inputs are $227 \times 227$ RGB patches randomly croppe from the resized images. Figure 2 illustrates the detailed architecture of our ConvNet. The convolution and max-pooling are two major operations in this network. The convolution operation can be expressed as:

$$y^j = \max(0, b^j + \sum_i k^{ij} * x^i) \quad (1)$$

where $x^i$ and $y^j$ respectively represent the $i$-th input map and $j$-th output map. $k^{ij}$ is the convolution kernel between the $i$-th input map and the $j$-th output map and $*$ denotes the convolution operation. A ReLU nonlinear activation function ($y = \max(0, x)$) is used after the convolution operation, which has been shown to achieve better fitting abilities than the sigmoid function [3]. The max-pooling operation can be formulated as:

$$y^i_{j,k} = \max_{0 \le m,n \le r}(x^i_{j \times r+m, k \times r+n}) \quad (2)$$

where each activation $y^i_{j,k}$ in the $i$-th pooling map $y^i$ pools over an $r \times r$ non-overlapping local region in the $i$-th input map $x^i$. The max-pooling operation can increase the receptive field of the neuron, while reducing the complexity of the network.

Let's indicate all the network parameters (including convolution and fully-connected layers) as $\theta$, which maps the input image $I_i$ to $x_i$: $x_i = f(I_i, \theta)$ where $x_i \in R^C$ with each element $x^j_i, j = 1, \dots C$ denotes the possibility that $I_i$ belongs to the $j^{th}$ place. The $x_i$ is then normalized at the final softmax layer to output a C-way softmax:

$$s_i \in R^C \text{ where } s^j_i = \frac{\exp(x^j_i)}{\sum_{k=1}^C \exp(x^k_i)}, j = 1, \dots, C. \quad (3)$$

The network is then learned by minimizing $-\log(s^t_i)$, with the $t^{th}$ target place. Stochastic gradient descent is used with gradients calculated using back-propagation. The batch size is set to 50.

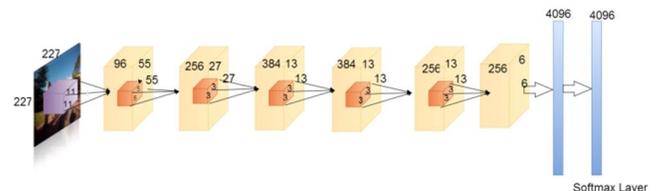

Figure 2 ConvNet Structure. The length, width, and height of each cuboid denotes the map number and the dimension of each map for all input and convolutional layers. The inside small cuboids denote the size of the convolution kernel. For example, the first cuboid after the input image indicates that there are 96 feature maps and each map is of size $55 \times 55$. The kernels between the input image and the first convolution layer are in size $11 \times 11$. The last two layers are fully-connected layers.

### B. Multi-scale Pooling

In this section, we employ a multi-scale pooling strategy inspired by [31] on the convolutional feature maps to further increase the features' robustness against viewpoint variations. Due to the parameter sharing mechanism of the convolution operation, a convolutional feature map can be interpreted as the detection scores obtained by applying the convolution filter on the input image and locations with high activation value indicate around them there are visual patterns that the filter is searching for. It is observed that generally a convolutional feature map is sparse in that only a few locations have high activations with the presence of certain visual patterns. This suggests that a convolution filter is highly selective to certain visual patterns. And when a place is observed from different angels, some of their visual patterns still preserve which can be detected by the same convolution filter. Figure 3 illustrates such examples.

With this observation, we apply a multi-scale pooling method to search for the most prominent visual patterns at multiple locations of the image in order to match images across different viewpoints. For each feature map, we apply a four-scale ($S = [1\ 2\ 3\ 4]$) pyramid pooling operation which first divides the image into $S \times S$ cells and within each spatial cell, we pool the responses using max pooling. The responses at the lowest scale ($S = 1$), for example, can be interpreted as the activations capturing the most salient structure within the whole image. After the pooling operation, a feature map with any size can be reduced to a vector of 30 dimensions which further reduces the computation complexity. In the end, we concatenate the pooling vectors from all the feature maps to form a representation of the image. In Section VI.B, we evaluate its effectiveness by comparing it with other state-of-the-art feature encoding methods.

### C. Implementation

Since the configurations of our top 5 convolution layers are similar to the CaffeNet [3], we can initialize our top 5 convolution layers using the weights learnt from the CaffeNet. This strategy is demonstrated to further improve the features's robustness. We call the CNN with weights initialized by the pre-trained CaffeNet as "HybridNet" and the one without such initialization as "AMOSNet". We compare their performances in Section VI.C and discuss their differences in Section VII.A. Both networks were trained using the Caffe package on a GPU NVIDIA Tesla K40. The initial learning rate is set to 0.01 which is divided by 10 after every 60,000 iterations. The weight decay is set to 0.005 with a momentum of 0.9. The training took about two days to finish a total of 120,000 iterations.

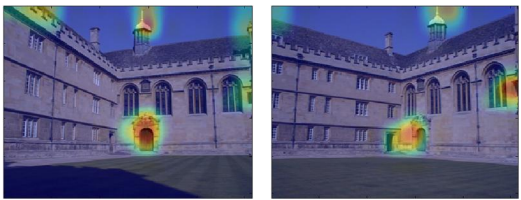

Figure 3 Convolutional feature maps from a pair of images captured under different viewpoints of the same place. It is observed that the feature maps are generally sparse and the same convolution filter fires at visually similar patterns even under different viewpoints.

### IV. CREATING THE SPECIFIC PLACES DATASET (SPED)

To create a sufficiently large place recognition dataset with enough environment variety and exemplars of each place under different conditions, we collected images captured from publicly accessible outdoor cameras, which have observed the same place over several years. We selected a subset of outdoor cameras from the Archive of Many Outdoor Scenes (AMOS) [32]. The AMOS consists of images captured from approximately 30,000 outdoor cameras around the world from March 2006 to 2017. Images from each camera were captured every half an hour for a period of ten years, which makes them ideal for studying long-term environment variations, such as season changes and lighting variations.

From 30,000 cameras, we randomly selected 2543 cameras and downloaded all the images captured by those cameras in February 2014 and August 2014, time points chosen that exhibit the strongest seasonal changes. In this way we constructed a dataset of about 2,500,000 images (and are still collecting more).

Downloaded images were curated. Firstly, we automatically removed images which were pitch black. These images were usually captured at night time in areas where there was no illumination. Then we removed corrupted images produced when the cameras were not functioning correctly. Some dataset examples are illustrated in Figure 4. Images in the database exhibit the following properties:

*1) Large environmental changes in each scene:*

For each camera, we collected all the images captured every half an hour in February 2014 and August 2014; this allows us to study environmental variations, such as lighting changes, day-night circles or season changes;

*2) Diversity across scenes*

The selected cameras cover a wide variety of outdoor scenes, ranging from forest landscapes, country roads to urban scenes;

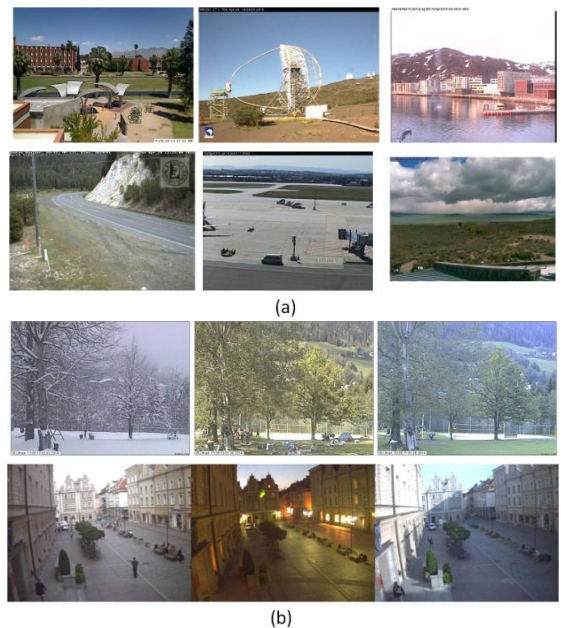

Figure 4 Sample images of the database. (a) Diversity across scenes; (b) Large condition variations in each scene; Each row represents images from the same place captured at different times;

### V. EXPERIMENTAL SETUP

In this section, we describe the testing datasets used and how we generate representations from images.

## A. Testing Datasets

The testing datasets are completely different to the training datasets; they come from geographically separated locations and encompass different types of environments. We tested using four benchmark place recognition datasets, with details summarized in Table I. Each dataset consists of several traverses along the same route under different conditions with one of the traverses used as a reference dataset and the other traverse used for testing.

The first three datasets were used to evaluate the robustness of the features against appearance changes. The Nordland dataset consists of 10 hour set video footage taken from the perspective of the front cart in four different seasons. The St. Lucia dataset was recorded in a suburb at different times of a day and contains medium viewpoint variations and significant appearance changes. The Eynsham dataset contains modest appearance variations.

The Gardens Point dataset is used to evaluate the viewpoint robustness of the features. It consists of two traverses along the same route. The first traverse was recorded at the daytime while keeping on the left side of the walkways, while the other traverse were taken at night from the right side of the walkways. As a result, this dataset exhibits both significant appearance and viewpoint changes.

## B. Ground Truth

For the Eynsham dataset, we used the 40 metre tolerance GPS-derived ground truth provided with the dataset, consistent with the tolerance used in the original FAB-MAP study [24]. Ground truth for both the Nordland and Gardens Point datasets was obtained by manually selecting frame correspondences. For the St. Lucia dataset, we first searched for all the reference images with the closest GPS coordinate and then visually matched the correct image.

TABLE I
DATASET DESCRIPTIONS

| Dataset Name | Number of Frames per Traverse | Environment |
| --- | --- | --- |
| Eynsham | 4789 | urban+suburban |
| St Lucia | 1000 | Suburban |
| Nordland | 1900 | Train journey |
| Garden Point | 200 | Campus |

## C. Feature Extraction

We compare feature activations from the third convolutional layer to the eighth fully connected layer. We use $L_k(I), k = 3, \ldots 8$ to denote the corresponding output of the $k^{th}$ layer given input image I. Each of these vectors is a deep learnt representation of the image I. For each layer output $L_k(I), k = 3, \ldots, 8$, we generate a corresponding confusion matrix $M_k, k = 3, \ldots, 8$ by matching each of the testing images to all reference frames. For example, each column $j$ in the confusion matrix stores the feature difference between the $j^{th}$ testing image to all reference frames.

## VI. RESULTS

In this section, we firstly visualize the weights and activations from different layers of the network, compare the performances of different encoding methods and then evaluate the individual robustness of the deep learnt features against the two main challenges in visual place recognition: appearance changes and viewpoint variations. All test datasets in this section were not used during the training of our model and therefore test the generalization power of the learnt features on unseen datasets.

### A. Visualization of the Deep Features

To gain a better understanding of what the networks have learned and the differences between our trained networks and other pre-trained CNNs, we visualize the weights and layer activations for different layers of the our model and CaffeNet. Visualizing the Conv 1 is straightforward in that we directly plot the weights in Conv 1. In the first row in Figure 6, we can see the weights of the first convolutional layer in both networks are very similar, both capturing the oriented edges and opponent colors of the images.

To visualize the units in higher layers, we first combine the test set of ImageNet LSVRC2015 (100,000 images) and SUN 397 (100,754 images) as the input for both networks; then we keep track of which images maximally activate some filters in each layer and visualize those images to get an understanding of what those filters are looking for in their receptive fields. In particular, as shown in Figure 6, we randomly pick four filters at each layer in layers 2,4,5 in both network, and show the top 9 image patches that maximally activate that filter, revealing the structures that excite that filter. It is shown that layer 2 in both networks respond to corners and other edge and color conjunctions. As the network propagates, larger differences can be observed at higher layers, indicating that features in higher layers start to capture the semantic meaning in the image. In particular, layer 4 and 5 of the CaffeNet are more interested in object-blobs, while the same layers in our model focuses more on shapes that look like landscapes with more spatial structures. The difference in these learned filter structures are closely relevant to the differences in the training data..

### B. Comparing Different Feature Encoding Methods

In this section, we compare the multi-scale pooling method utilized in Section III.B with two other state-of-the-art convolution feature encoding methods – the cross-layer pooling [33] and the holistic pooling [34]. Both the multi-scale and holistic pooling are performed on the last convolutional layer (layer 6), while the cross-layer pooling is evaluated using the last two convolutional layers (layer 5 and 6). As shown in Figure 5, despite its simplicity, the multi-scale pooling method consistently achieves better performances than the other two methods, indicating this pooling method may be more suitable for place recognition.

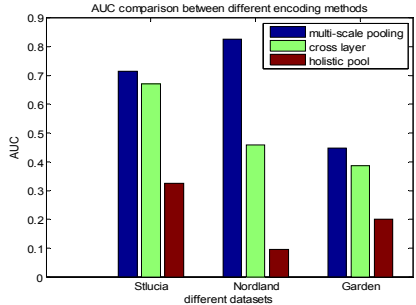

Figure 5 Performance comparison between different encoding methods on the Stlucia, Nordland and Garden datasets

## C. Benchmark Place Recognition Performance

In this section, we evaluate the place recognition performance of different network layers. We present Area Under the curve (AUC) on three test datasets by using the activations from different layers of the networks as generic features of the images. No feature encoding method is used here. Particularly, we compare the features generated by the AMOSNet, HybridNet, CaffeNet [3] and PlaceNet [12]. We also compare our network with the SeqSLAM [16] which is a state-of-the-art place recognition algorithm. The 'Conv6' layer in Figure 7 to 10 refers to the sixth convolutional layer for AMOSNet and HybridNet and the sixth fully-connected layer for CaffeNet and PlaceNet.

Figure 7 present AUC generated by the AMOSNet, HybridNet, CaffeNet and PlaceNet on the Nordland dataset. The AMOSNet and HybridNet consistently outperform SeqSLAM, CaffeNet and PlaceNet on all the layers. The advantage is most obvious on the fifth convolutional layer and the best performance is achieved by using the fifth convolutional layer of HybridNet. We suspect this is because images variations with each camera in the SPED dataset contain, despite its environmental dynamics, are smaller than those in each class in ImageNet. Since AMOSNet is only trained on SPED, it learns less discriminative features than HybridNet, which combines the discriminative power of CaffeNet fine-tune specifically for place recognition. Also noteworthy is that convolutional layers uniformly perform better than fully-connected layers, partly due to the fact that spatial information is retained in the convolutional layers. This result is also consistent with the image retrieval experiments of [10, 35] which suggest that the middle network layers provide a more general feature description while the top layers are overtrained for the ImageNet and place recognition tasks respectively.

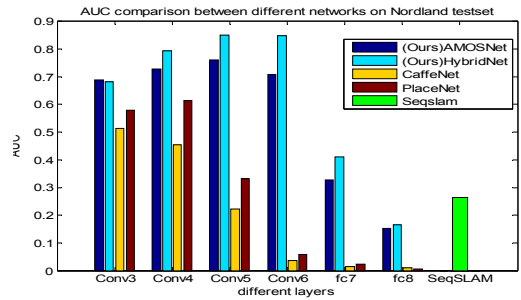

Figure 7. AUC comparing the performances of features extracted from the AMOSNet (blue), HybridNet (light blue), CaffeNet (yellow) and PlaceNet (brown) on the Nordland dataset.

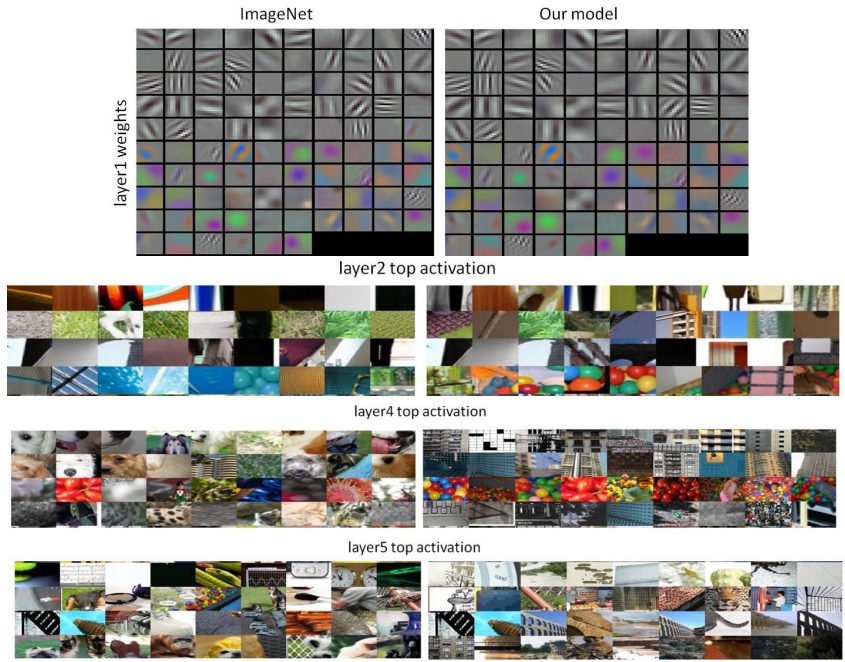

Figure 6. Visualization of the weights from the first convolutional layer (top row) and top 9 activations at other higher layers for the ImageNet (left column) and our models (right column). For layer 2-5, we show the top 9 activations for a random subset of filters at each layer across a validation data of 208,754 images. Although weights and activation patches at early layers are similar, we observe semantically different activation patches at higher layers between these two networks. For example, at layer 4, the second row, the filter from the ImageNet (left) fires more on object-blob, such as the face of a dog, while our model (right) fires more on scene-type patches, such as the buildings.

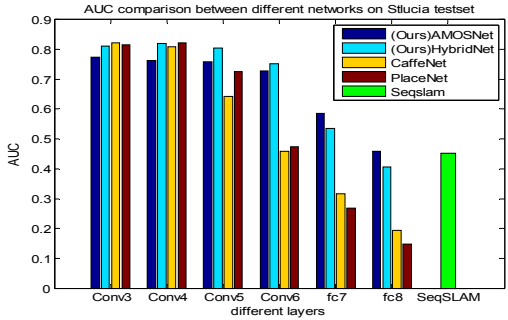

Figure 8. AUC comparing the performances of features extracted from the AMOSNet (blue), HybridNet (light blue), CaffeNet (yellow) and PlaceNet (brown) on the Stlucia dataset..

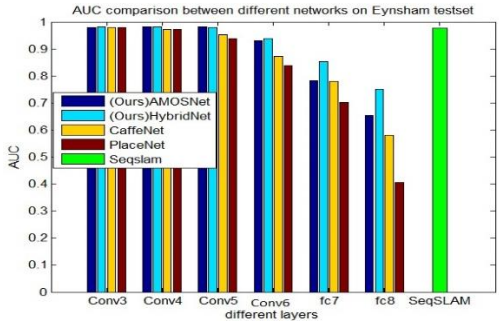

Figure 9. AUC comparing the performances of features extracted from the AMOSNet (blue), HybridNet (light blue), CaffeNet (yellow) and PlaceNet (brown) on the Eynsham dataset. Our models: the HybridNet (light blue) and AMOSNet (blue) achieves better performance than other CNNs in layers Conv6, fc7 and fc8.

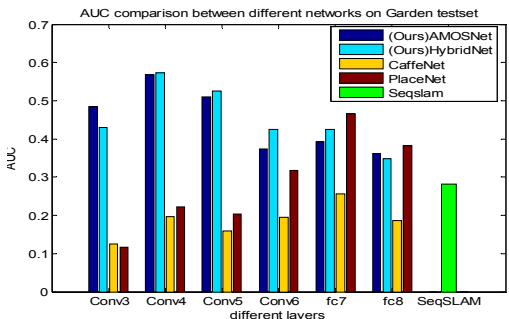

Figure 10 AUC comparing the performances of features extracted from the AMOSNet (blue), HybridNet (light blue), CaffeNet (yellow) and PlaceNet (brown) on the Gardens Point dataset. Our models: the HybridNet (light blue) and AMOSNet (blue) achieves better performance than other CNNs in layers Conv3, Conv4, Conv5 and Conv6

Figure 8 present the results on the St. Lucia dataset. All four networks perform comparably in the third and fourth convolutional layers. However, in the other layers such as Conv5, Conv6, fc7 and fc8, the AMOSNet and HybridNet uniformly generate better performance than the CaffeNet and PlaceNet, illustrating its robustness against illumination variations.

Figure 9 describe the results on the Eynsham dataset. It is clear that both networks perform very well on this dataset, probably because there are not significant appearance changes between the testing and reference traverses. It is worthy noted that the best performance is still achieved by the HybridNet on the Conv5 layer, and the AMOSNet and HybridNet still outperform the CaffeNet and PlaceNet on all the network layers.

### D. Viewpoint Change Robustness

In this section, we evaluate the learnt features' robustness against viewpoint changes on the Gardens Point dataset, although the training dataset did not explicitly contain exemplars of viewpoint change. The Gardens Point dataset exhibits both appearance and viewpoint changes.

Figure 10 illustrates that both the AMOSNet and HybridNet produce features that are more robust against appearance and viewpoint variations than CaffeNet on all the network layers. The HybridNet produces the most robust features in layers Conv4, Conv5 and Conv6 while the PlaceNet outperforms all other networks in layers fc7 and fc8. The best performance is observed in layer Conv4, produced by the HybridNet.

## VII. DISCUSSION AND FUTURE WORK

In this paper, we presented CNN training process utilizing a new, very large-scale condition-invariant place recognition database with millions of images from a wide variety of environments, to study how appearance changes over time. We trained two deep CNNs on this large-scale dataset and demonstrated on several challenging datasets that the internal representations learned from the networks are more robust against appearance and viewpoint variations than those extracted from object-centric networks, such as CaffeNet. We compare our network with state-of-the-art place recognition algorithms and demonstrate its superior performance. We also provided a visualization of the weights and layer activations of the CNN units to illustrate the differences in the internal representation between networks learned from place-centric and object-centric databases. Future work will pursue a number of promising avenues of investigation.

### A. HybridNet vs. AMOSNet

In the experiments, we observe that HybridNet consistently achieves better performance than AMOSNet which is counterintuitive. We expected that the AMOSNet train from scratch should be more place recognition-specific and achieve better performance. One possible reason is that although the whole training dataset contains over one million images, the variations within each camera is relatively small compared to the variations observed in ImageNet; therefore, training a network completely from scratch on this dataset may end up not learning enough useful structures. The HybridNet, since it is fine-tuned from CaffeNet, can carry useful structure that has been learned from the ImageNet and is therefore more discriminative.

### B. More significant viewpoint invariance

The current network is trained on a dataset collected from static webcams, with little viewpoint variation. The challenge faced in this research is that there is no dataset with significant viewpoint changes which is large enough to train a deep CNN. We are currently working on generating large scale (city-size) high fidelity synthetic datasets with an unlimited range of viewpoint and condition variation in order to train networks from scratch to be condition *and* viewpoint invariant.

Perhaps one of the most exciting future avenue of research revolves around the increasing availability (if shared) of extremely large quantities of place-relevant data obtained from self-driving car fleets being operated by the major corporates and start-ups. With appropriate curation and pre-processing, these datasets may enable us to train place recognition systems that significantly surpass even the current state of the art results. However, even with this new car-based data, it is likely that traditional large scale datasets such as the one we have developed here will remain relevant, if not for any other reason that they represent a broader range of application domains where place recognition and navigation remain critical robotic competencies.